\documentclass[sigconf]{acmart}

\AtBeginDocument{%
  }

\setcopyright{acmlicensed}
\copyrightyear{2018}
\acmYear{2018}
\acmDOI{XXXXXXX.XXXXXXX}

\acmConference[CIKM '2024]{The 33rd ACM International Conference on 
Information and Knowledge Management}{October 21-25,
  2024}{Boise, Idaho, USA.}
\acmISBN{978-1-4503-XXXX-X/18/06}

\begin{document}

\title{MODRL-TA:A Multi-Objective Deep Reinforcement Learning Framework for Traffic Allocation in E-Commerce Search}

% \author{Anonymous Authors}
\author{Peng Cheng} 
\email{chengpeng58@jd.com}
\author{Huimu Wang}
 \authornote{corresponding author}
\email{wanghuimu1@jd.com}
\author{Jinyuan Zhao}
\email{zhaojinyuan1@jd.com}
\author{Yihao Wang} 
\email{wangyihao8@jd.com}
\affiliation{%
  \institution{JD.com}
  \city{Beijing}
  \country{China}
}

\author{Enqiang Xu}
\email{xuenqiang@jd.com}
\author{Yu Zhao}
 \authornote{corresponding author}
\email{zhaoyu384@jd.com}
\author{Zhuojian Xiao}
\email{xiaozhuojian5@jd.com}
\author{Songlin Wang}
\email{wangsonglin3@jd.com}
\affiliation{%
  \institution{JD.com}
  \city{Beijing}
  \country{China}}

\author{Guoyu Tang}
\email{tangguoyu@jd.com}
\author{Lin Liu}
\email{liulin1@jd.com}
\author{Sulong Xu} 
\email{xusulong@jd.com}
\affiliation{%
  \institution{JD.com}
  \city{Beijing}
  \country{China}}

\renewcommand{\shortauthors}{Anonymous Authors}

\begin{abstract}
  Traffic allocation is a process of redistributing natural traffic to products by adjusting their positions in the post-search phase, aimed at effectively fostering merchant growth, precisely meeting customer demands, and ensuring the maximization of interests across various parties within e-commerce platforms. Existing methods based on learning to rank neglect the long-term value of traffic allocation, whereas approaches of reinforcement learning suffer from balancing multiple objectives and the difficulties of cold starts within real-world data environments. To address the aforementioned issues, this paper propose a multi-objective deep reinforcement learning framework consisting of multi-objective Q-learning (MOQ), a decision fusion algorithm (DFM) based on the cross-entropy method(CEM), and a progressive data augmentation system(PDA). Specifically. MOQ constructs ensemble RL models, each dedicated to an objective, such as click-through rate, conversion rate, etc. These models individually determine the position of items as actions, aiming to estimate the long-term value of multiple objectives from an individual perspective. Then we employ DFM to dynamically adjust weights among objectives to maximize long-term value, addressing temporal dynamics in objective preferences in e-commerce scenarios. Initially, PDA trained MOQ with simulated data from offline logs. As experiments progressed, it strategically integrated real user interaction data, ultimately replacing the simulated dataset to alleviate distributional shifts and the cold start problem. Experimental results on real-world online e-commerce systems demonstrate the significant improvements of MODRL-TA, and we have successfully deployed MODRL-TA on an e-commerce search platform. 
\end{abstract}


\received{20 February 2007}
\received[revised]{12 March 2009}
\received[accepted]{5 June 2009}

\maketitle

\section{Introduction}
Traffic allocation plays a crucial role within E-commerce platforms, primarily through algorithmic optimization to adjust organic traffic distribution, in contrast to traffic from advertising. Positioned after the re-ranking phase in the search sorting pipeline, it seeks to achieve specific business objectives by changing product positions. These objectives include enhancing new product visibility; swiftly introducing potential bestsellers to target groups during holidays or specific periods; and dynamically adjusting promotional strategies based on market feedback, aiming to maximize benefits for the platform, merchants, and consumers.

There are existing heuristic method for achieving the traffic allocation \cite{chen2012ad,bharadwaj2012shale,aastrom2006pid}. However, these methods solely focus on a single item’s benefits, which overlook the reality that the allocation strategy change of one item might affect the optimal strategies of other items. Therefore, great efforts have been made on developing reinforcement
learning-based techniques \cite{cai2017real,wang2018learning,rohde2018recogym,wu2018multi,zhao2021dear,wang2021hierarchical,xie2021hierarchical}, which can continuously update their advertising strategies during the interactions with consumers, and the optimal strategy is made by maximizing the expected long-term cumulative revenue from consumers. 

However, most existing works focus on maximizing the single utility of items, while ignoring the multiple utility of items and merchants like conversion rate, click rate or add-cart rate. Multi-objective reinforcement learning methods \cite{nguyen2020multi,abels2019dynamic,mossalam2016multi} can make the balance between multiple objectives by multi-reward shaping or ensemble learning while their optimal strategy are relatively static. Specifically, the business objectives of merchants change dynamically over time. For instance, For a merchant newly joining the platform, user clicks are actually more important than orders (to cultivate the mindset of attracting users to browse), but after a period of time, the merchant will pay more attention to orders and GMV. Multi-objective reinforcement learning emphasizes finding a static Pareto optimum, whereas in our scenario, there is a need for real-time dynamic adjustment of objective weights. Furthermore, the aforementioned reinforcement learning methods encounter cold start problems due to the sparsity of real data in the early stages of online deployment.

To track these aforementioned issues, a multi-objective deep reinforcement learning framework (MODRL-TA) is proposed in this paper.
Specifically, it consists of three components including a multi-objective Q-learning (MOQ), a real-time decision fusion module (DFM) based on the Cross-Entropy Method (CEM)\cite{de2005tutorial}, and a progressive data augmentation system (PDA). Specifically, the MOQ optimizes different objectives (such as traffic completion, click-through rate, conversion rate) by using multiple reinforcement learning models. Its advantages include: first, precise optimization for each objective; second, easy scalability with new objectives through adding models without retraining existing ones; third, flexible integration of model outputs for optimal decision-making based on changing business needs. Upon completion of the model training, the DFM utilizes the CEM to find a set of suitable weight parameters, aiming to maximize the joint expectation of all models under the same action while meeting business constraints. Finally, for the cold-start of MODRL, the PDA calculates the initial state from offline logs, selects actions randomly, and then calculates the new state and reward, repeating this process to generate multiple sets of state, action, and reward data. This simulated data is used for the model's cold start phase. Upon deployment, the model is initially trained with 100\% simulated data. Following the launch, it gradually accumulates real data. Training then proceeds with a mix of 90\% simulated data and 10\% real data, gradually increasing the proportion of real data until the optimal training outcome is achieved.

The contributions of this paper can be summarized as follows:
\begin{itemize}
\item[$\bullet$] We propose a multi-agent reinforcement learning framework that achieves precise modeling and real-time optimization of multiple objectives under dynamically changing business constraints to maximize joint profits. 
\item[$\bullet$] The PDA in MODRL-TA mitigates the cold start problem by initially training the MOQ model with simulated data from offline logs and strategically incorporating real user interaction data as experiments progress, eventually replacing the simulated dataset entirely. 
\item[$\bullet$] We conduct extensive experiments on a real-world dataset. Experimental results show that our model achieves significant
improvement compared with existing methods. MODRL-TA has been successfully deployed on the home search platform in a large E-commerce platforms and brought substantial economic benefits.

\end{itemize}

\vspace{-2.0em}
\section{Problem Formulation}
In this paper, we study the allocation problem within search list as a Markov Decision Process (MDP)\cite{van2012reinforcement}. The MDP consists of a tuple
of five elements $(S,A,P,R,\gamma)$: At each time step $t$, the agent observes the state $s_t \in S$, chooses an action $a_t \in A$ according to the policy $\pi$, gets a reward $r_t \in R$, and transitions to next state $s_{t+1}$ according to transition probability $p(s_{t+1}|s_t,a_t)$. The objective is to maximize the total expected discounted reward $R_{t}= {\textstyle \sum_{t=0}^{T}}\gamma ^{t}r^{t}$, where $\gamma \in [0,1]$ is a discount factor. We conceptualize the allocation system as the agent, where the action corresponds to the positioning of items displayed in the search results. The users serve as the environment, with their feedback acting as the reward signal.

\section{Methods}
In this section, we presenter the overview structure of MODRL-TA in Figure 1. 
To be more specific, considering the precision, scalability and flexible integration, MOQ constructs multiple separate RL models for multiple objectives instead of constructing a single RL model with multiple objectives. Upon completion of the RL model training, the DFM utilizes the CEM to find a set of suitable weight parameters, aiming to maximize the joint expectation of all models. Finally, PDA utilized a method of progressive sampling with both offline and online data to solve the cold start problem of RL.

\begin{figure*}[h]
  \centering
  \includegraphics[scale=0.35]{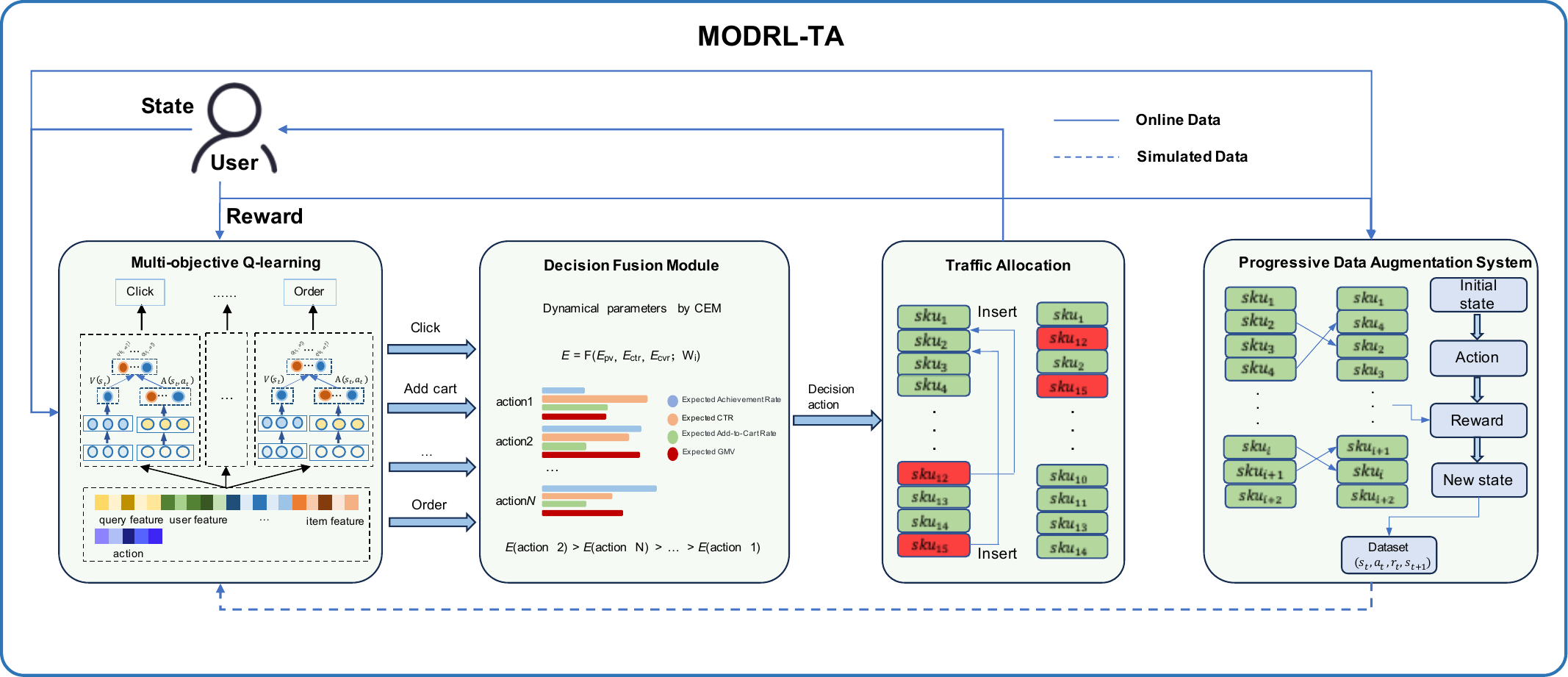}
  \caption{The framework of MODRL-TA}
  \label{tod1}
  \vspace{-2.0em}
\end{figure*}

\subsection{Multi-objective Q-learning}
As aforementioned, the traffic allocation problem is challenging because (i) It should maximize the long-term value of multiple objectives;(ii) The multiple business objectives of merchants change dynamically over time; (iii) Merchants often have new optimization goal requirements added. To address these challenges, we constructs multiple separate RL models combined with DFM instead of using traditional MORL. MORL emphasizes finding a static Pareto optimum, whereas in our scenario, there is a need for real-time dynamic adjustment of objective weights. In addition to meeting the needs for dynamic weight adjustment, separate RL models also demonstrate excellent scalability. When merchant present new objective requirements, there is no need to retrain existing models. Instead, training a new model specifically for the new objectives suffices, significantly enhancing training efficiency. Without loss of generality, the classic DQN\cite{mnih2013playing} has been selected as the foundational RL model for MODRL-TA in this paper.

\subsubsection{States and Actions}
The states $s_t$ in our reinforcement learning model mainly include the following parts:
\begin{itemize}
    \item User profile features, including gender, age group, etc.
    \item Query attribute features, including intent classification, etc.
    \item User historical behavior features, including items the user has clicked on (added to cart/ordered), item categories, etc.
    \item Contextual features, including item-related features.
    \item User new items historical behavior features, including new products the user has clicked on, etc.
    \item Total feedback features for control objectives.
\end{itemize}
These features include both fixed attribute features of the user and query, user sequence preference features obtained by modeling user historical behavior, and features of the item list displayed under the current request. This allows RL model to integrate information from the user, query, and item, thereby achieving personalized modeling at the request level. 

For the action space, $a_t \in \mathbb{R}^L$ is the location of item to interpolate the selected items (given a list of L items)

\subsubsection{Reward Design}
Since we are using multi-objective reinforcement
learning, we utilize multiple reinforcement learning models,
each controlling a specific objective. In this paper, we discuss strategies for optimizing objectives, using clicks and orders as examples. Specifically, Model A is dedicated to enhancing click-through rates, while Model B focuses on increasing the number of orders. The difference between these reinforcement learning models lies in their reward design. For the Model A controlling clicks, we set the reward of a state to 1 if a user clicks on a specific item, and -1 if there is no click. The reward design for the model controlling orders is similar, i.e.,
\begin{equation}
    \text{$r_t$} = 
    \begin{cases}
        1, & \text{if Click/Order} \\
        -1, & \text{if Only Page View}
    \end{cases}
\end{equation}

% \subsubsection{Action Space}
% Since our reinforcement learning model operates at the request level and needs to score all SKUs under a single request for each position, our action space corresponds to the ranking positions. For instance, if our candidate ranking positions are the top 100, then our action space consists of 100 discrete values from 0 to 99, corresponding to the positions from the 1st to the 100th. The optimal position is the best action chosen by the model:

% \begin{equation}
%     a^* = \arg\max_a g^*(s_{t+1}, a)
% \end{equation}

\subsubsection{Training Tasks}
The Deep Q-network, i.e., action-value function of the $i-th$ RL model $Q^i(s_t, a_t)$, can be optimized by minimizing a sequence of loss functions $L^i(\theta^i)$ as:
\begin{equation}
L^i(\theta ) = \mathbb{E}_{s^i_t,a^i_t,r^i_t,s_{t+1}}(y^i_t-Q^i(s_t^i,a_t^i;\theta^i ))^{2} 
\end{equation}
where $y_t^i = \mathbb{E}_{s^i_{t+1}}[r_t+\gamma max_{a^i_{t+1}}Q^i(s^i_{t+1},a^i_{t+1};\theta^i_T|s^i_t,a^i_t )]$ is the target for the current iteration of the $i-th$ model. We introduce separated evaluation and target networks \cite{mnih2013playing} to help
smooth the learning and avoid the divergence of parameters,
where $\theta^i$ represents all parameters of the evaluation network,
and the parameters of the target network $\theta^i_T$ are fixed when optimizing the loss function $L^i(\theta)$. The derivatives of loss function $L^i(\theta)$ with respective to parameters 
\begin{equation}
\bigtriangledown _\theta^i L^i(\theta^i ) = \mathbb{E}_{s_t^i,a_t^i,r_t^i,s_{t+1}^i}(y_t^i-Q^i(s_t^i,a_t^i;\theta^i )) \bigtriangledown_{\theta^i}Q(s_t^i,a_t^i;\theta^i )
\end{equation}
where $y_t^i = \mathbb{E}_{s^i_{t+1}}[r_t+\gamma max_{a^i_{t+1}}Q(s^i_{t+1},a^i_{t+1};\theta^i_T|s_t^i,a_t^i )]$  and $max_{a^i_{t+1}}$ will look through the location of item to interpolate.

In this paper, all RL models share inputs and jointly update the parameters of the input layer, while the remaining parameters are updated independently, which allows the models to share knowledge during the learning process, and the total loss is: $L_{total}=\sum_{0}^{I} L^i(\theta )$

\subsection{Decision Fusion Module}
Upon modeling the long-term expectation of different objectives, the next step is to consider how to balance these objectives to meet the dynamically changing needs of merchants. The DFM employ the Cross Entropy Method (CEM)\cite{de2005tutorial} as the balance strategy. CEM is an evolutionary algorithm that excels in optimizing parameters in complex, high-dimensional spaces. Its simplicity and adaptability make it ideal for various optimization problems. In e-commerce, CEM efficiently handles dynamic traffic and immediate feedback adjustments, quickly tuning parameters in response to changing merchant feedback and overcoming cold start issues with new products or users through initial interactions.

The main idea of CEM is to maintain a distribution of potentially optimal solutions, and then update this distribution based on the samples and the sampled values obtained from querying the black box. Treat \(w \) as a vector, where \(n\) samples are taken in each round:
\begin{equation}
    w_i \sim \mathcal{N}(\mu_t, \sigma_t^2), \quad i \in [n]
\end{equation}
where $\mu_t$ represents the mean, and $\sigma_t^2$ represents the variance.

Then, perform a rollout to obtain the corresponding values \(S(w_1), S(w_2), \ldots, S(w_n)\). Select the top \(n\) values, obtaining their index set \(I \subseteq [n]\), and then update the distribution:

\begin{equation}
    \mu_{t+1} = \frac{\sum_{i \in I} w_i}{|I|}
\end{equation}

\begin{equation}
    \sigma_{t+1}^2 = \frac{\sum_{i \in I} (w_i - \mu_{t+1})^T (w_i - \mu_{t+1})}{|I|} + Z_{t+1}
\end{equation}
where \(Z_{t+1}\) is a noise to prevent the variance from converging too quickly.

% After finding the optimal parameters, when dealing with tasks targeting two objectives, the final gain is fused linearly:

% \begin{equation}
%     \text{gain}(s_t, a) = w_1 \cdot g_{\text{click}}(s_t, a) + w_2 \cdot g_{\text{order}}(s_t, a)
% \end{equation}

% Calculate the AUC performance for different tasks based on the fused \(\text{gain}(s_t, a)\), the following takes clicks as an example.:

% \begin{equation}
%     \text{auc}_{\text{click}} = \text{cal}_{\text{auc}}(\text{label}_{\text{click}}, \text{gain}(s_t, a))
% \end{equation}

% For different scenarios, the tuning objective also varies. The goal is to maximize the target task's AUC as a constraint condition for parameter iteration.

To address multi-objective tasks in e-commerce, we propose a gain fusion method that linearly combines the gains of different objectives. Similar to Section 3.1, we use the optimization objectives of clicks and orders as examples. Specifically, we define the combined gain function as:
\begin{equation}
    \text{gain}(s_t, a) = w_1 \cdot q_{\text{click}}(s_t, a) + w_2 \cdot q_{\text{order}}(s_t, a),
\end{equation}
where $q_{\text{click}}(s_t, a)$ represents the Q-value of model controlling clicks and $q_{\text{order}}(s_t, a)$ represents the Q-value of model controlling orders. The weights $w_1$ and $w_2$ are tunable parameters that balance the importance of clicks and orders based on business priorities. Moreover, the weights are subjected to a performance metric based on $\text{gain}(s_t, a)$. For instance, the rea Under ROC (AUC) can be used as the performance metric for different tasks: 
\begin{equation}
    \text{auc}_{\text{click}} = \text{cal}\_{\text{auc}}(\text{label}_{\text{click}}, \text{gain}(s_t, a)). 
\end{equation}

Our goal is to maximize the AUC of the target task during parameter iteration to achieve optimal performance. This method allows us to dynamically adjust and optimize the trade-off between different objectives, ensuring that the system remains responsive to varying merchant needs and market conditions.

\subsection{Progressive Data Augmentation System}
During the system cold start phase, since there is no online real data available for RL model training, we design a progressive data augmentation system as shown in Firure\ref{tod1} to obtain training data for the cold start phase. Specifically, we adjust the display positions of candidate items, then calculate the position changes of all candidate items. The pCTR (predicted Click-Through Rate) will change after the item positions are adjusted. When there is a conflict in the positions after adjusting the candidate item positions, we set a higher priority for the item originally ranked higher, and the item with a lower priority is moved backward.

The pCTR after item position adjustment is calculated by first determining the CTR corresponding to each display position over a past period. After adjusting the item positions, the pCTR also adjusts accordingly: if a item is moved from index \(j\) to index \(i\), the CTR change rate after the position adjustment is calculated. This change rate, multiplied by the original pCTR of the item, gives the pCTR after the position adjustment:
\begin{equation}
    \text{pCTR}' = \text{pCTR} \times \left( \frac{\text{CTR}_{\text{index}_i}}{\text{CTR}_{\text{index}_j}} \right)
\end{equation}
After obtaining the cold start training data using offline simulation data, the trained model is deployed online, allowing the accumulation of real online data. As the amount of real data increases, we gradually replace the offline model data with real online data. After a period, the real data will cover 100\% of the training data.

\section{Experiments}
\subsection{Datasets and Metrics}
To evaluate the proposed method, we compiled an internal e-commerce dataset, including all sessions with purchasing behaviors over the past 15 days, based on users' historical queries. The dataset, organized by user request timing to depict state transitions, contains about 4 million records. We measure effectiveness using CTR (Click-Through Rate) and CVR (Conversion Rate)~\cite{xu2024optimizing, wang2024preference, adaptive}, common e-commerce metrics.

% In order to accurately evaluate the effectiveness of the proposed method, we have compiled an internal dataset from an e-commerce scenario. This dataset encompasses all sessions that include purchasing behaviors within the past 15 days, as identified from users' historical queries, and is organized according to the timing of user requests to facilitate a precise depiction of the transition process of users' state behaviors. The entire dataset comprises approximately 4 million records. To measure the metrics, we use CTR (Click-Through Rate) and CVR (Conversion Rate), which are commonly used indicators in the e-commerce scenario.

\subsection{Experimental Settings }
\subsubsection{Methods}
Our baseline model is the PID algorithm\cite{johnson2005pid}, a common feedback control method that adjusts the system's input to ensure the output follows the desired outcome quickly and accurately. However, since PID lacks a reward metric, we use the multi-objective reinforcement learning with fusion reward (MORL-FR) deployed online as our baseline. This model initially improved revenue online, and we iterated offline to maximize rewards. To evaluate performance, we use cumulative rewards, focusing on CTR and CVR. Additionally, we conducted ablation experiments with models trained on offline simulated data, real online data, without the CEM module, and the latest model version.

% Our baseline model is the PID algorithm, a common feedback control method that adjusts the control system's input to ensure the output follows the desired output quickly and accurately.

% To evaluate the performance of different models, we use cumulative rewards as a metric, specifically measuring CTR and CVR. Since the PID algorithm lacks a reward concept, after achieving initial revenue with single-objective reinforcement learning online, we iterated offline with the goal of maximizing rewards. Additionally, we conducted several ablation experiments, including models trained entirely with offline simulated data, models trained entirely with real online data, models without the CEM module, and the latest version of the model.

\subsubsection{Implementation Details}
The replay memory has a 2,000,000 capacity with a warmup size for pre-loading data. A discount factor of 0.999 prioritizes future rewards. The target network updates every 200 steps for stability. The action space includes 100 decisions, and state features span 222 dimensions. 

% The replay memory is set at a capacity of 2,000,000 to store past transitions for future training, complemented by a warmup size for pre-loading data. The discount factor, at 0.999, prioritizes future rewards. Updates to the target network occur every 200 steps to ensure training stability. The action space encompasses 100 possible decisions, and the state features are captured in 222 dimensions, covering all relevant inputs. These hyperparameters are fine-tuned based on specific application needs and performance feedback, aiming for optimal algorithm performance.

\subsection{Offline Experimental Results}
Starting with MORL-FR as a baseline, which achieves a CTR Reward of 5.88 and a CVR Reward of 0.63, the table showcases incremental improvements through different configurations of the MODRL-TA algorithm. Notably, the MODRL-TA with 100\% Simulated Data improves upon the baseline, indicating the framework's capability to enhance performance even without real-world data. The version with 100\% Real Data maintains the CTR Reward but significantly boosts the CVR Reward to 0.97, highlighting the importance of real data in increasing conversion efficiency. A remarkable leap is observed with MODRL-TA without the CEM, suggesting that even in the absence of CEM, the framework substantially outperforms the baseline. The fully optimized MODRL-TA configuration achieves the highest performance, with a CTR Reward of 12.20 and a CVR Reward of 2.25, demonstrating the comprehensive effectiveness of the MODRL-TA framework, including CEM, in optimizing e-commerce metrics. This progression underscores the potential of advanced reinforcement learning techniques in refining both engagement and conversion outcomes in digital platforms.

\begin{figure}
    \centering
    \includegraphics[width=1\linewidth]{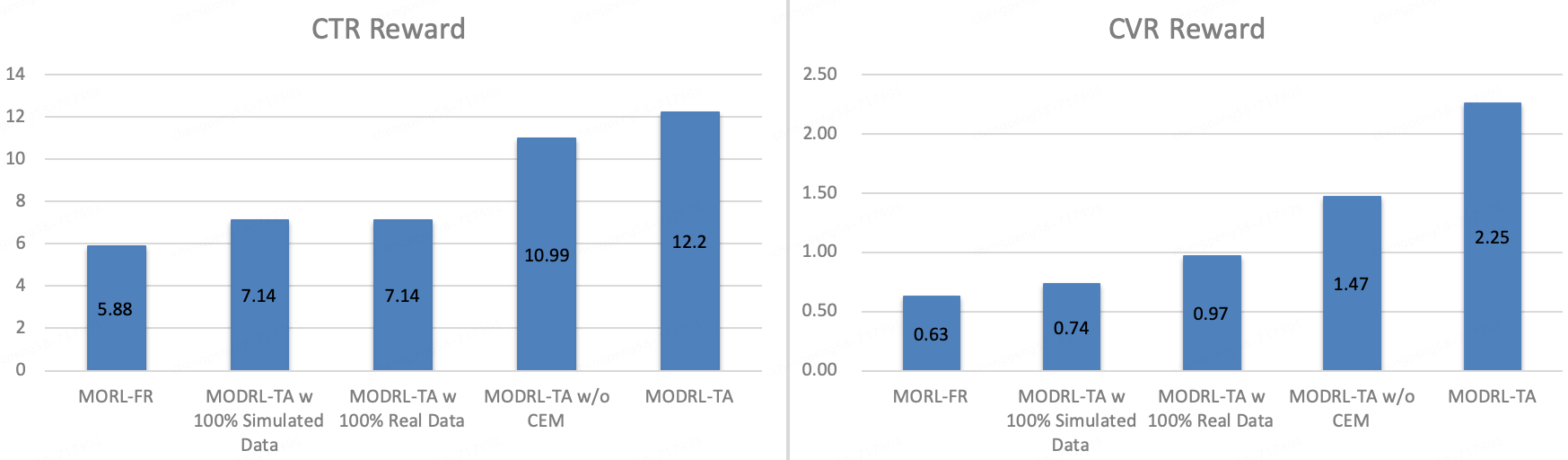}
    \caption{Model Performance Comparison}
    \vspace{-0.4cm}
    \label{fig:enter-label}
\end{figure}

\begin{table}[htbp]
    \vspace{-1.0em}
    \centering
    \caption{Model Performance Comparison}
    \label{table1}
    \begin{tabular}{c|c c}  
        \hline
        Algorithm & CTR Reward & CVR Reward \\  
        \hline
        MORL-FR & 5.88 & 0.63 \\
        MODRL-TA w 100\% Simulated Data & 7.14 & 0.74 \\
        MODRL-TA w 100\% Real Data & 7.14 & 0.97 \\
        MODRL-TA w/o CEM & 10.99 & 1.47 \\
        MODRL-TA  & 12.20 & 2.25 \\
        \hline
    \end{tabular}
    \vspace{-1.5em}
\end{table}

\subsection{Online Experimental Results}
After achieving benefits in offline evaluations, we conducted a two-week online A/B test. Compared to the PID algorithm, The MODRL-TA, particularly when integrating the Request Dimension, dramatically improves performance, showing an increase in Impressions (IMP) by up to +18.0\%, in Click-Through Rate (CTR) by up to +4.2\%, and in Conversion Rate (CVR) by up to +5.1\%. These results indicate a significant leap in performance across all metrics when transitioning from MORL to MODRL-TA optimization and further incorporating real data and request dimensions into the learning process.
This model has already been successfully deployed online, serving 570 million daily active users.
% \begin{table}[htbp]
%     \centering
%     \vspace{-1.0em}
%     \caption{Model Performance Comparison}
%     \label{table1}
%     \setlength{\tabcolsep}{1pt} % 设置列间距为1pt
%     \begin{tabular}{c|c c c c}  
%         \hline
%         Algorithm & IMP & CTR  & CVR  & p-value\\  
%         \hline
%         Single-objective RL & +3.39\% & +1.88\% & +0.92\%  & 0.000\\
%         MODRL-TA w 100\% Simulated Data & +8.44\% & +3.80\% & +2.94\% & 0.001\\
%         MODRL-TA w 100\% Real Data & +13.23\% & +3.93\% & +3.12\% & 0.007\\
%         MODRL-TA w Request Dimension & +18.02\% & +4.25\% & +5.14\% & 0.000\\
%         \hline
%     \end{tabular}
%     \vspace{-1.5em}
% \end{table}

\begin{table}[htbp]
    \centering
    \vspace{-1.0em}
    \caption{Model Performance Comparison}
    \label{table1}
    \setlength{\tabcolsep}{1pt} % 设置列间距为1pt
    \begin{tabular}{c|c c c c}  
        \hline
        Algorithm & IMP & CTR  & CVR  & p-value\\  
        \hline
        MORL-FR & +3.3\% & +1.8\% & +0.9\%  & 0.000\\
        MODRL-TA w 100\% Simulated Data & +8.4\% & +3.8\% & +2.9\% & 0.001\\
        MODRL-TA w 100\% Real Data & +13.2\% & +3.9\% & +3.1\% & 0.007\\
        MODRL-TA w Request Dimension & +18.0\% & +4.2\% & +5.1\% & 0.000\\
        \hline
    \end{tabular}
    \vspace{-1.5em}
\end{table}

% \begin{table}[htbp]
%     \centering
%     \vspace{-1.0em}
%     \caption{Model Performance Comparison}
%     \label{table1}
%     \setlength{\tabcolsep}{2pt} % 设置列间距为1pt
%     \begin{tabular}{c|c c c c}  
%         \hline
%         Algorithm & IMP & CTR  & CVR  & p-value\\  
%         \hline
%         Single-objective RL & 3.3\% & 1.8\% & 0.9\%  & 0.000\\
%         MODRL-TA w 100\% Simulated Data & 8.4\% & 3.8\% & 2.9\% & 0.001\\
%         MODRL-TA w 100\% Real Data & 13.2\% & 3.9\% & 3.1\% & 0.007\\
%         MODRL-TA w Request Dimension & 18.0\% & 4.2\% & 5.1\% & 0.000\\
%         \hline
%     \end{tabular}
%     \vspace{-1.5em}
% \end{table}

\section{Conclusions}
This paper proposes a multi-objective deep reinforcement learning framework (MODRL-TA) for balancing dynamically changing multiple objectives in traffic allocation. It consists of three components including MOQ, DFM and PDA. MOQ optimizes different objectives by using multiple reinforcement learning models with sharing inputs. DFM utilizes the CEM to find a set of suitable weight parameters, aiming to maximize the joint expectation of all models with dynamically changing multiple objectives. PDA mitigates the cold start problem by progressively integrating simulated and online data. Extensive experiments conducted on a real-world dataset and the successful deployment on an e-commerce search platform show that the proposed framework can achieve significant benefits.

\bibliographystyle{ACM-Reference-Format}
\bibliography{sample-base}

% \appendix









\end{document}